\documentclass[lettersize,journal]{IEEEtran}
\usepackage{amsmath,amsfonts}
\usepackage{algorithmic}
\usepackage{algorithm}
\usepackage{array}
\usepackage[caption=false,font=normalsize,labelfont=sf,textfont=sf]{subfig}
\usepackage{textcomp}
\usepackage{stfloats}
\usepackage{url}
\usepackage{verbatim}
\usepackage{graphicx}
\usepackage{cite}
\usepackage{booktabs}
\usepackage{gensymb}
\usepackage{cases}
\usepackage{amssymb}
\makeatletter

\newcommand{\Rmnum}[1]{\expandafter\@slowromancap\romannumeral #1@}
\makeatother
\newcommand{\XNAME}{TactV}

\begin{document}

\title{\XNAME : A Class of Hybrid Terrestrial/Aerial \protect\\ Coaxial Tilt-Rotor Vehicles}

\author{Yifei Dong, Yimin Zhu, Lixian Zhang~\IEEEmembership{Fellow,~IEEE}, Yihang Ding
\thanks{(Yifei Dong and Yimin Zhu are co-first authors.) (Corresponding author: Lixian Zhang.)}
\thanks{The authors are with the School of Astronautics, Harbin Institute of Technology, Harbin 150080, China (e-mail: yfdong@stu.hit.edu.cn; ymzhu@stu.hit.edu.cn; yhding@stu.hit.edu.cn; lixianzhang@hit.edu.cn).}}

\markboth{Journal of \LaTeX\ Class Files,~Vol.~14, No.~8, August~2021}%
{Shell \MakeLowercase{\textit{et al.}}: A Sample Article Using IEEEtran.cls for IEEE Journals}


\maketitle

\begin{abstract}
    To enhance the obstacle-crossing and endurance capabilities of vehicles operating in complex environments, this paper presents the design of a hybrid terrestrial/aerial coaxial tilt-rotor vehicle, \XNAME, which integrates advantages such as lightweight construction and high maneuverability. Unlike existing tandem dual-rotor vehicles, \XNAME~employs a tiltable coaxial dual-rotor design and features a spherical cage structure that encases the body, allowing for omnidirectional movement while further reducing its overall dimensions. To enable \XNAME~to maneuver flexibly in aerial, planar, and inclined surfaces, we established corresponding dynamic and control models for each mode. Additionally, we leveraged \XNAME's tiltable center of gravity to design energy-saving and high-mobility modes for ground operations, thereby further enhancing its endurance. Experimental designs for both aerial and ground tests corroborated the superiority of \XNAME's movement capabilities and control strategies. 
\end{abstract}


\section{Introduction}

\IEEEPARstart{R}{esearch} on hybrid terrestrial/aerial vehicles has made significant progress in recent years, leading to widespread applications in rescue exploration, indoor mapping, military reconnaissance, and payload transportation \cite{ref1,ref2,ref3,ref4,ref5}. Due to the complex and variable nature of actual application scenarios, these vehicles often require prolonged mission execution and agile maneuverability in their environments. Consequently, achieving high maneuverability while maintaining low energy consumption has become a focal point of research in hybrid terrestrial/aerial vehicles.

The energy efficiency of aerial movement in these vehicles is directly related to the number and configuration of rotors. According to actuator disk theory \cite{ref6}, with the same load, a larger rotor disk area leads to lower loading per unit area, thereby increasing hover efficiency. Therefore, with a constant rotor disk area, reducing the number of rotors allows for a corresponding decrease in the overall size of the the unmanned aerial vehicle (UAV). Thus, dual-rotor configurations exhibit higher energy efficiency compared to other multirotor designs. Numerous research teams have explored various aspects of dual-rotor designs. Tandem tiltable dual-rotor vehicles, such as those in studies \cite{ref7,ref8,ref9,ref10}, etc. utilize vector thrust to control the airframe. A common feature of these dual-rotor vehicles is that they control rotor tilt via servo motors to achieve attitude control. Similarly, Qin et al.\cite{ref11} proposed a design scheme of a dual rolls swashplate-less mechanism. Instead of using servo motors to achieve vector thrust, they controls attitude by using cyclic flapping response in hinges that connect the blades, so that the number of actuators is reduced. It is evident that existing dual-rotor structures predominantly adopt tandem layouts. However, coaxial dual-rotor configurations can achieve higher energy efficiency under the same length and width constraints.

Many existing coaxial dual-rotor vehicles \cite{ref12,ref13,ref14,ref15,ref16} are often structurally redundant and complex, significantly increasing their volume and mass, resulting in unnecessary energy consumption. This paper proposes a miniaturized and lightweight design for coaxial dual-rotor vehicles. The size of our design is comparable to existing rotor vehicles, yet it significantly enhances energy efficiency. We employ two servos to control the tilting of the coaxial motors in the pitch and roll directions, generating thrust vectoring. Additionally, we achieve yaw control through differential speed from the two rotors, ensuring that the vehicle's attitude during aerial movement is entirely controllable.

\begin{figure}[t]
  \centering
  \includegraphics[width=1.0\columnwidth]{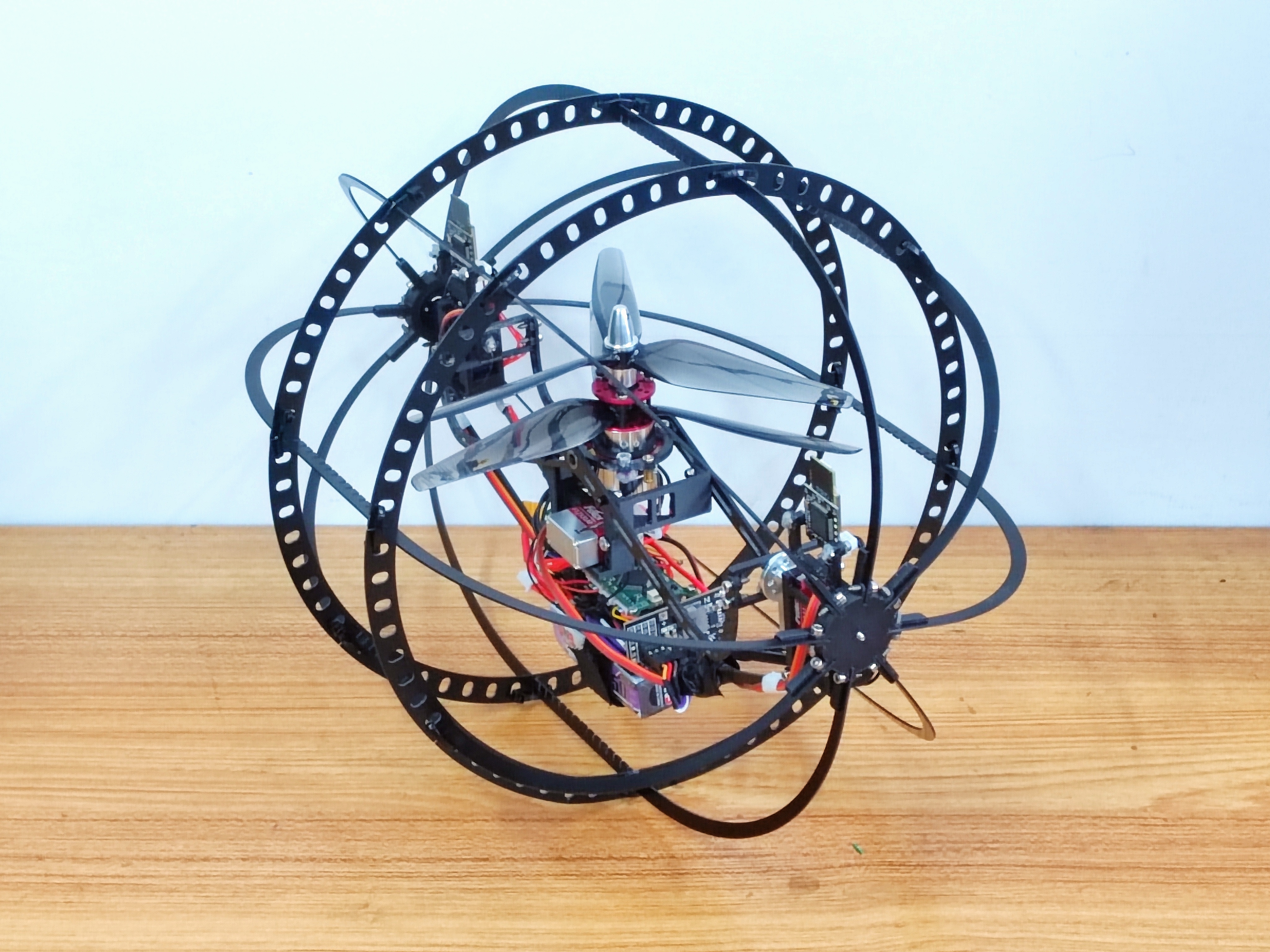}
  \caption{\XNAME~mechanical design detail and avionics}
  \label{fig_3}
\end{figure}

The movement modes of hybrid terrestrial/aerial vehicles on the ground have been extensively studied. Recent research indicates that wheeled and cage-like designs effectively balance performance and energy consumption, making them quite common. Pan et al.\cite{ref17} outfitted a quadrotor with a passive omnidirectional wheel, controlling its ground attitude and movement through the thrust generated by the propellers. This approach provides significant flexibility on the ground but increases energy consumption during terrestrial mode. Studies \cite{ref18} and \cite{ref19} implemented two passive wheels on either side of the vehicle, similarly controlling ground movement through aerodynamic forces. This design improves stability and energy efficiency during ground operations. Yang et al.\cite{ref10} utilized a spherical cage surrounding the rotors, merging protective and ground mobility functions. In contrast to previous designs, this vehicle only requires tilting the rotors to achieve ground movement, further reducing energy consumption during terrestrial operations. In summary, despite the design of wheels or spherical cages reducing the energy consumption of the UAV during ground movement, its propulsion is primarily provided by the thrust generated from rotor rotation. If the vehicle can leverage the tilt of its center of gravity to provide ground movement propulsion, energy consumption can be further minimized.

\begin{figure}[t]
  \centering
  \includegraphics[width=1.0\columnwidth]{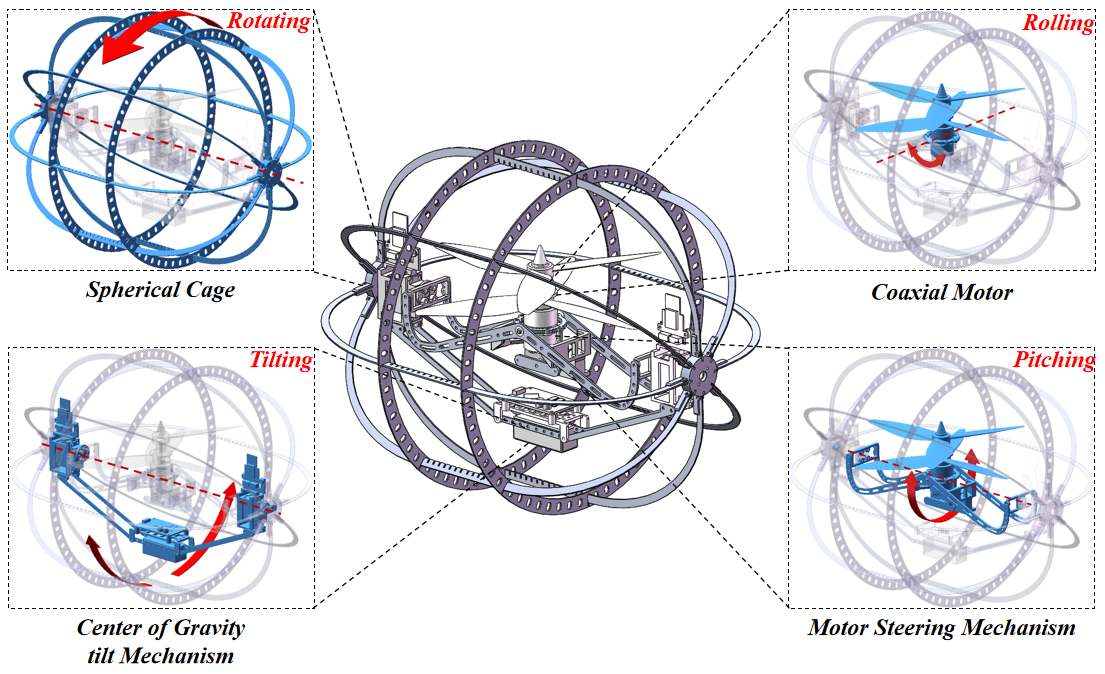}
  \caption{The movement of the institutions on \XNAME.}
  \label{fig_4_0}
\end{figure}

Providing propulsion through center of gravity tilt is commonly seen in spherical robots \cite{ref20,ref21,ref22,ref23}. This paper integrates this approach with rotor-based vehicles. We encase the vehicle in a spherical cage and use a servo motor to control the relative movement between the internal body and the external cage. The spherical cage is constructed from fiberglass (FR-4), resulting in low mass. Consequently, when the body tilts forward, the shift in the center of gravity generates torque, propelling the vehicle forward. This method reduces energy consumption compared to solely relying on aerodynamic push. Besides focusing on low-power design, we also investigate ground maneuverability. When the vehicle requires significant acceleration or is moving along an incline, the center of gravity tilt may not provide sufficient forward acceleration. Thus, we combine the thrust with center of gravity tilt to enhance ground maneuverability at the expense of energy consumption. These two ground propulsion modes can be switched freely based on actual needs.

This paper primarily introduces the overall structure, motion modeling, controller design, ground movement strategies, and motion performance of the coaxial dual-rotor vehicle. The main contributions of this paper are as follows:
\begin{itemize}
  \item[1)]
We innovatively propose a hybrid terrestrial/aerial vehicle, named \textbf{\XNAME} (hybrid \textbf{T}errestrial/\textbf{a}erial \textbf{c}oaxial \textbf{t}ilt-rotor \textbf{V}ehicles). By controlling the pitch and roll of the coaxial motors and the differential speed of the two rotors, \XNAME~can achieve omnidirectional movement in both aerial and terrestrial environments. The lightweight materials used in the design of the spherical cage ensure that \XNAME~remains highly maneuverable while maintaining high strength and ventilation.

  \item[2)]
We analyze \XNAME's motion modes in aerial, planar, and inclined scenarios, establishing corresponding dynamic and control models based on its structure.

  \item[3)]
We propose energy-saving and high-mobility modes for \XNAME's ground movement and design corresponding controllers. The integration of center of gravity tilt with aerodynamic thrust further reduces energy consumption and enhances endurance.

\end{itemize}

The remainder of this paper is organized as follows: Section \Rmnum{2} describes the mechanical design details of \XNAME. Section \Rmnum{3} analyzes various motion modes of \XNAME~in the air and on the ground, along with dynamic modeling. Section \Rmnum{4} discusses the system's multimodal control and energy consumption analysis. Section \Rmnum{5} presents and interprets the results of \XNAME's motion experiments. Finally, Section \Rmnum{6} concludes the paper.

\section{Mechanical Design}

Fig.\ref{fig_3} illustrates the mechanical design of \XNAME, which includes the coaxial tilt-rotor system, the associated steering mechanism, the center of gravity tilt structure, and the spherical cage structure. The motion mechanisms of these components are depicted in Fig.\ref{fig_4_0}. Under servo control, the coaxial motors can roll, the motor steering mechanism can pitch, the center of gravity tilt mechanism can tilt, and the spherical cage structure can roatate. It is important to note that the tilting of the center of gravity structure and the rolling of the spherical cage are controlled by the same servo motor, and the principles behind this will be explained in subsequent sections.

\subsection{Motor Steering Mechanism}

\begin{figure}[t]
\centering
\includegraphics[width=1.0\columnwidth]{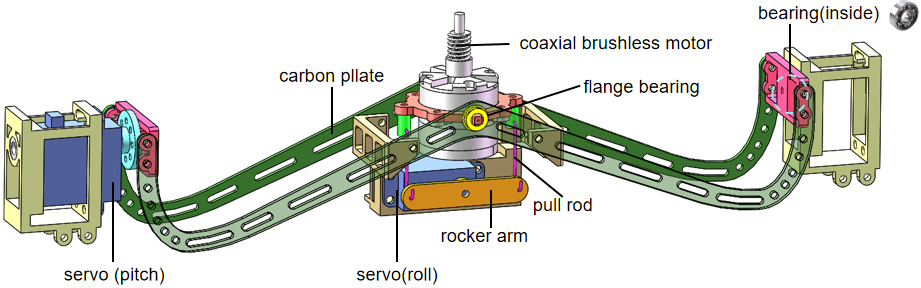}
\caption{The main components of motor steering mechanism}
\label{fig_4}
\end{figure}

Fig.\ref{fig_4} shows the motor steering mechanism, which conceals a pair of GWS7035 rotors. The CR23M coaxial brushless motor ($1550KV$) is equipped with two flange bearings mounted at the front and back, which are embedded in the holes reserved in the carbon plates. The FT2331M servo (roll) located below controls the roll movement of the motor through a connected rocker arm and pull rod. The  servo (pitch) is connected to the two carbon plates, controlling the overall pitch movement of the central mechanism. A ball bearing is embedded at the same position on the opposite side of the carbon plate as the servo (pitch), reducing rotational damping and ensuring the symmetry and continuity of the pitch movement.

As shown in Fig.\ref{fig_4}, the carbon plate design of the motor steering mechanism is shaped like a ``$W$'' to prevent collisions between the rotors and other mechanisms during rotation around the roll axis. Additionally, when designing the rotating axis, it is essential to align the intersection of the pitch and roll axes with the center of gravity of the motor steering mechanism (not to be confused with \XNAME's overall center of gravity) to avoid influencing \XNAME's center of gravity during the rotation of the motor steering mechanism.

\subsection{Center of Gravity Tilt Mechanism}

\begin{figure}[t]
\centering
\includegraphics[width=1.0\columnwidth]{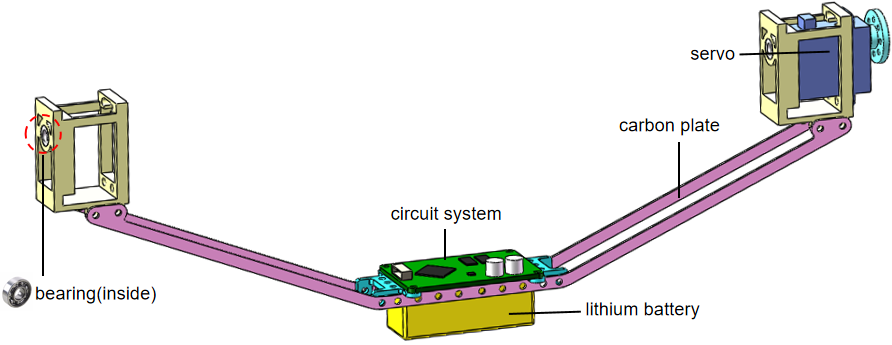}
\caption{The main components of the center of gravity tilt mechanism}
\label{fig_5}
\end{figure}

The center of gravity tilt mechanism is illustrated in Fig.\ref{fig_5}. This mechanism is located below the motor steering mechanism. The circuit system, together with a $3S$, $850mAh$ lithium battery used as a weight, is fixed at the lowest point of the carbon plate. The carbon plate is connected to an FT2331M servo, with the servo's disc fixed to the outer spherical cage structure. In terrestrial mode, the servo controls the angle of tilt for the weight, allowing for changes in \XNAME's acceleration; the specific modeling method will be detailed in Section \Rmnum{3}.

\subsection{Spherical Cage Structure}

\begin{figure}[t]
  \centering
  \includegraphics[width=0.8\columnwidth]{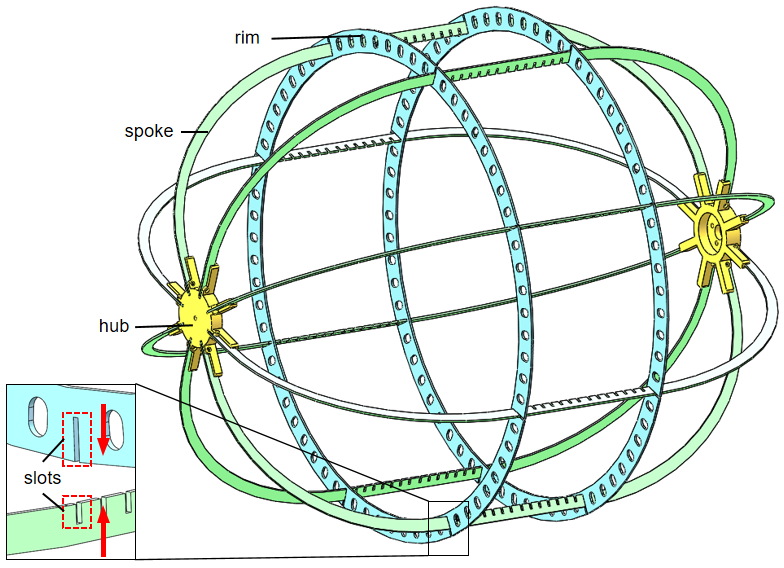}
  \caption{The main components of the spherical cage structure, along with the combination of the rims and spokes.}
  \label{fig_6}
\end{figure}

The spherical cage structure is depicted in Fig.\ref{fig_6}. The hubs at both ends of the cage are made from PA12 nylon via 3D printing, connecting eight spokes made of FR-4. To prevent instability caused by movement between the spokes, two FR-4 rims are added for reinforcement. The outer side of the rims has slots corresponding to the inner side of the spokes, and after the spokes and rims are assembled, zip ties are used for additional support. The spherical cage structure weighs approximately $60g$, accounting for about $20\%$ of the total weight.

\subsection{Circuit System}

\XNAME~is currently equipped with various sensors to obtain real-time attitude data and environmental information. These include an BMI088 for attitude detection and a receiver for remote control signals. Additionally, \XNAME~is equipped with two ultra wide band(UWB) modules for bidirectional communication with the ground station, transmitting current and desired positional and attitudinal information. Currently, \XNAME~maintains a hover throttle of approximately $55\%$, indicating substantial payload capacity.

\subsection{Wind Resistance}

\begin{figure}[t]
\centering
\includegraphics[width=1.0\columnwidth]{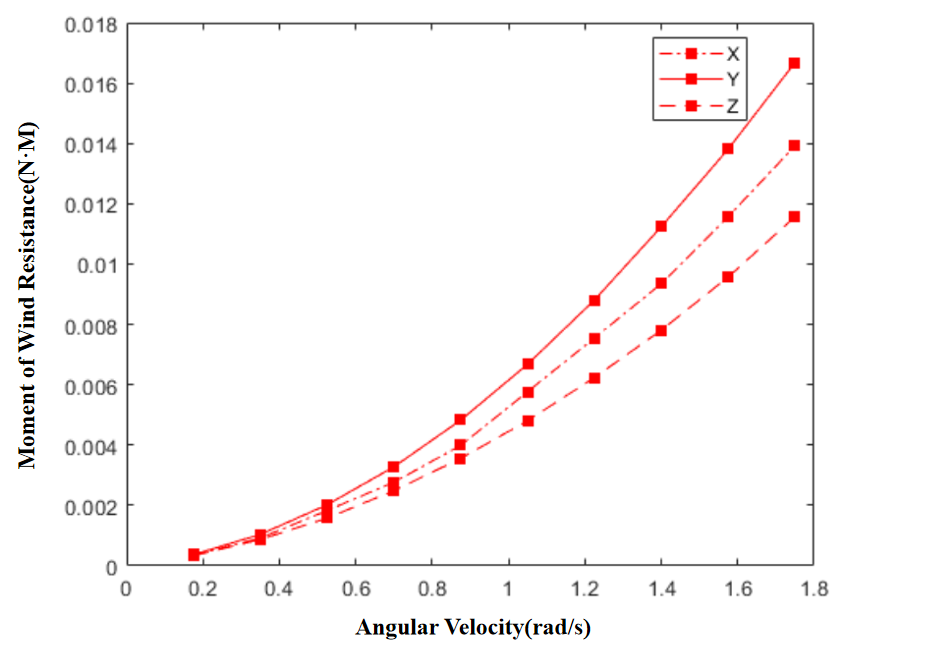}
\caption{Relation between the angular velocity and the aerodynamic torque of \XNAME}
\label{fig_7}
\end{figure}

Wind resistance is a crucial factor affecting \XNAME's performance, consisting of aerodynamic drag and torque. Reducing wind resistance during movement not only extends endurance but also enhances agility and maneuverability. The formula for calculating aerodynamic drag is given by:
\begin{equation}
\label{eq1}
  f_{A} = 0.5C_{d} \rho \boldsymbol{v}^{2} A
\end{equation}
\noindent where $C_{d}$ is the drag coefficient, $\rho$ is the air density, $\boldsymbol{v}$ is the relative velocity between the object and the air, and $A$ is the frontal area. The calculation of torque due to wind resistance is more complex, primarily depending on the frontal area and the angular velocity of the body. Therefore, in the design process, we focus on minimizing the frontal area to reduce drag. Computational fluid dynamics simulations are conducted to analyze the airflow around \XNAME~by varying its linear and angular velocities to obtain the corresponding aerodynamic drag and torque. The relationship between aerodynamic torque and angular velocity is illustrated in Fig.\ref{fig_7}, showing that the aerodynamic torque exhibits a quadratic relationship with angular velocity.

\section{Dynamics Modeling}

\begin{figure*}[t]
  \centering
  \subfloat[]{\includegraphics[width=2.3in]{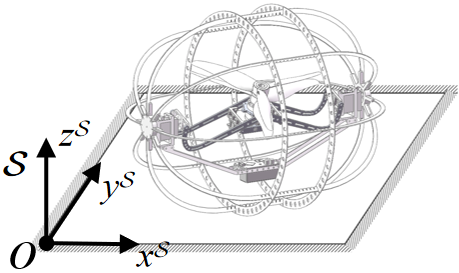}%
  \label{fig_8_1}}
  \hfil
  \subfloat[]{\includegraphics[width=2.3in]{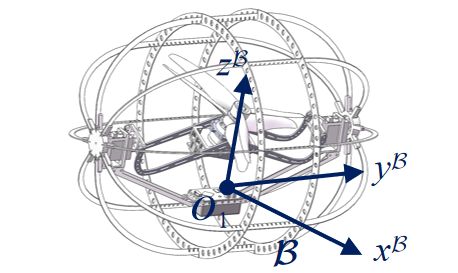}%
  \label{fig_8_2}}
  \hfil
  \subfloat[]{\includegraphics[width=2.3in]{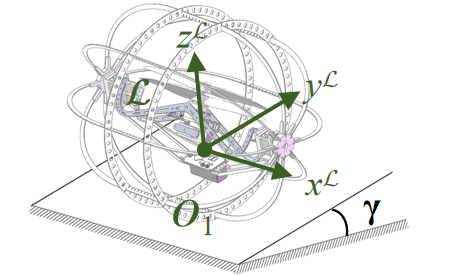}%
  \label{fig_8_3}}
  \caption{Definition of coordinate frames of \XNAME. (a) The ENU inertial frame $\mathcal{S}$. (b) The body frame $\mathcal{B}$. (c) The slope frame $\mathcal{L}$.}
  \label{fig_8}
\end{figure*}

The motion of \XNAME~can be classified into various modes based on environmental conditions and operational requirements. From the perspective of drive modes, it is divided into aerial and terrestrial modes. In terms of motion strategy, it can be categorized into energy-efficient and maneuverability-prioritized modes. To gain a deeper understanding of the distinctions between different motion modes, this section analyzes \XNAME's movement in various scenarios.

Before modeling, it is crucial to establish appropriate coordinate systems. The three right-handed coordinate systems of \XNAME~are illustrated in Fig.\ref{fig_8}. The East-North-Up (ENU) inertial frame $\mathcal{S}$ serves as the world coordinate system, with its origin coinciding with the origin $O$ of the test site, and its $z$-axis oriented opposite to the direction of gravity. The origin of the body frame $\mathcal{B}$ aligns with the center of gravity tilt mechanism's center $O_{1}$, with the $z$-axis normal to the upper surface of the battery and the $y$-axis matching the direction of the rotation axis shown in Fig.\ref{fig_8}. The angles $\varphi$, $\theta$ and $\psi$ denote the pitch, roll, and yaw angles of $\mathcal{B}$ relative to $\mathcal{S}$. The angles $\alpha$ and $\beta$ represent the pitch and roll angles of the brushless motor relative to $\mathcal{B}$. The slope frame $\mathcal{L}$ has its origin at the center of gravity $O_{1}$ of \XNAME, with the $z$-axis aligned with the normal direction of the slope and the $x$-axis directed along \XNAME's forward motion on the slope. The slope angle is denoted as $\gamma$.

Based on the aforementioned coordinate systems, we will conduct a mechanical analysis and dynamic modeling of \XNAME's motion in both aerial and terrestrial modes.

\subsection{Aerial Mode}

\XNAME's aerial movement resembles that of conventional tilt-rotor configurations, exhibiting six degrees of freedom, including all translational and rotational freedoms. Notably, \XNAME's yaw motion is uniquely achieved through torque generated by differential rotor speeds. The formula for calculating rotor torque is given by
\begin{equation}
\label{eq2}
  \tau _{p}=C_{p}\rho (n_{1}^{2}-n_{2}^{2})D^{5} 
\end{equation}
\noindent where $C_{p}$ is the rotor torque coefficient, $n_{1}$ and $n_{2}$ are the rotational speeds of the two rotors, and $D$ is the diameter of the paddle disc.

The translational and rotational dynamics of \XNAME~in aerial mode can be expressed using the Newton-Euler equations as follows
\begin{subequations}
  \begin{align}
    &M\boldsymbol{a}^{\mathcal{S}} = \boldsymbol{G}^{\mathcal{S}}+\boldsymbol{R}^{\mathcal{BS}A}(\boldsymbol{f}^{\mathcal{B}}_{a}+\boldsymbol{F}^{\mathcal{B}})\label{eq3}\\
    &\boldsymbol{J}^{\mathcal{B}}\dot{\boldsymbol{\omega}}^{\mathcal{B}} + \boldsymbol{\omega}^{\mathcal{B}}\times \boldsymbol{J}^{\mathcal{B}}\boldsymbol{\omega}^{\mathcal{B}} = \boldsymbol{\tau} _{fa}^{\mathcal{B}} +\boldsymbol{\tau} _{F}^{\mathcal{B}}+\boldsymbol{\tau} _{p}^{\mathcal{B}} \label{eq4}
  \end{align}
\end{subequations}
\noindent where $M$ is \XNAME's total mass, $\boldsymbol{a}^{\mathcal{S}}$ is \XNAME's acceleration (with the superscript $\mathcal{S}$ denoting the $\mathcal{S}$ frame, the same below), $\boldsymbol{G}^{\mathcal{S}}=\left [ 0\;0\;-\!Mg \right ]^{T}$ is the weight, $\boldsymbol{R}^{\mathcal{BS}A}$ is the rotation matrix (with the superscript $\mathcal{BS}$ indicating the rotation matrix from $\mathcal{B}$ to $\mathcal{S}$, the same below), $\boldsymbol{f}^{\mathcal{B}}_{a}=\left [ -f_{ax}\;-\!f_{ay}\;-\!f_{az} \right ]^{T}$ is the aerodynamic drag, and $\boldsymbol{F}^{\mathcal{B}}=\left [ Fcos\alpha sin\beta\;-\!Fsin\alpha\;Fcos\alpha cos\beta \right ]^{T}$ is thrust. The thrust can be calculated using the formula
\begin{equation}\label{eq6}
  F=C_{T}\rho(n^2_1+n^2_2)
\end{equation}
\noindent where $C_{T}$ is the thrust coefficient, $\boldsymbol{J}^{\mathcal{B}}$ is the inertia matrix. The inertia matrix $\boldsymbol{J}^{\mathcal{B}}$ can be simplified based on computations in analysis software.
\begin{equation}\label{J}
  \boldsymbol{J}^{\mathcal{B}}=
  \begin{bmatrix}
    J_{xx} & 0 &0 \\
    0 & J_{yy}& 0\\
    0 & 0 &J_{zz} 
   \end{bmatrix}
\end{equation}
\noindent$\dot{\boldsymbol{\omega}}$ and $\boldsymbol{\omega}$ are the angular acceleration and angular velocity of \XNAME, respectively. $\boldsymbol{\tau}^{\mathcal{B}} _{fa}=\left [ -\tau_{fax}\;-\!\tau_{fay}\;-\!\tau_{faz} \right ]^{T}$ is the wind resistance moment, and $\boldsymbol{\tau}^{\mathcal{B}} _{F}$ is the torque of thrust. Equations (\ref{eq3}) and (\ref{eq4}) can be used to obtain the translational dynamics equation of mass center of \XNAME~in the $\mathcal{S}$ frame and the angular dynamics equation in the $\mathcal{B}$ frame
\begin{small}
  \begin{subequations}
    \begin{align}
      &M\ddot{x}=F[S_{\alpha}(C_{\varphi}C_{\psi}-S_{\varphi}C_{\psi}S_{\theta})+C_{\alpha}C_{\beta}(S_{\varphi}S_{\psi}\notag\\
      &\quad\quad\;\;\; +C_{\varphi}C_{\psi}S_{\theta})+C_{\alpha}S_{\beta}C_{\psi}C_{\theta}]-f_{Ax}\label{eq5_1}\\
      &M\ddot{y}=-F[S_{\alpha}(C_{\varphi}C_{\psi}+S_{\varphi}S_{\psi}S_{\theta})+C_{\alpha}C_{\beta}(C_{\varphi}C_{\psi}\notag\\
      &\quad\quad\;\;\; -C_{\varphi}S_{\psi}S_{\theta})-C_{\alpha}S_{\beta}S_{\psi}C_{\theta}]-f_{Ay}\label{eq5_2}\\
      &M\ddot{z}=-Mg-F(C_{\alpha}S_{\beta}S_{\theta}+S_{\alpha}S_{\varphi}C_{\theta}-C_{\alpha}C_{\beta}C_{\varphi}C_{\theta})\notag\\
      &\quad\quad\;\;\; -f_{Az}\label{eq5_3}\\
      &J_{xx}\ddot{\varphi}=[-\tau_{fx}+\tau_{p}C_{\alpha}S_{\beta}+(J_{yy}-J_{zz})\dot{\theta}\dot{\psi}+FlC_{\alpha}S_{\beta}]\label{eq5_4}\\
      &J_{yy}\ddot{\theta}=[\tau_{fy}-\tau_{p}S_{\alpha}-(J_{xx}-J_{zz})\dot{\varphi}\dot{\psi}-FlS_{\alpha}]\label{eq5_5}\\
      &J_{zz}\ddot{\psi}=[-\tau_{fz}+\tau_{p}C_{\alpha}C_{\beta}-(J_{yy}-J_{xx})\dot{\varphi}\dot{\theta}]\label{eq5_6}
    \end{align}
  \end{subequations}
\end{small}
\!\!\!\noindent where $S_{\alpha}$, $S_{\alpha}$, $S_{\beta}$, $S_{\varphi}$, $S_{\theta}$, $S_{\psi}$ are shorthands for the sine function, and $C_{\alpha}$, $C_{\alpha}$, $C_{\beta}$, $C_{\varphi}$, $C_{\theta}$, $C_{\psi}$ are shorthands for the cosine function. $l$ represents the distance between $O_{1}$ and $O_{2}$. 

\subsection{Planar Mode}

\begin{figure}[t]
  \centering
  \subfloat[]{\includegraphics[width=0.49\columnwidth]{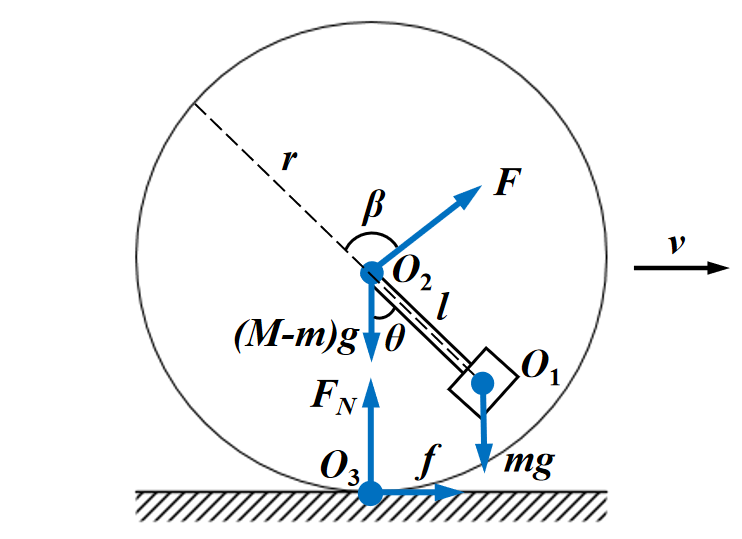}%
  \label{fig_9}}
  \subfloat[]{\includegraphics[width=0.49\columnwidth]{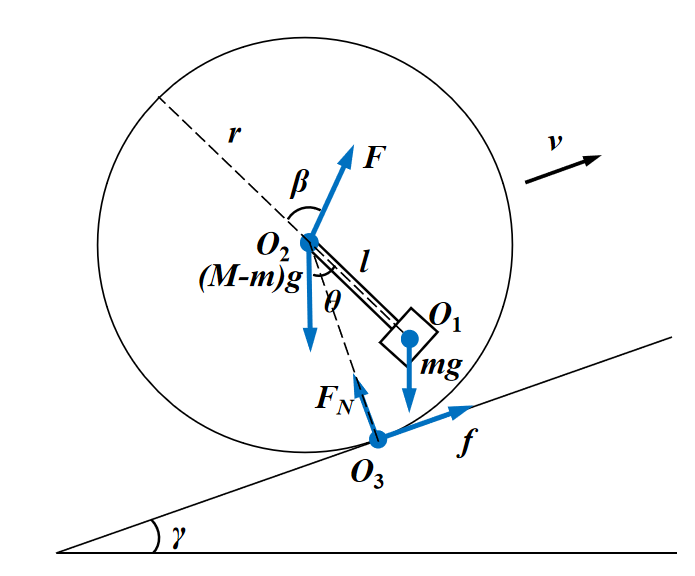}%
  \label{fig_10}}
  \caption{The force analysis of \XNAME~, where (a) is planar mode and (b) is inclined mode.}
  \label{fig_9_10}
\end{figure}

\XNAME's terrestrial mode encompasses planar motion (planar mode) and slope motion (inclined mode). In terrestrial mode, the dynamics involve torque generated by center of gravity shifts as well as thrust produced by rotor rotation. The contribution of these two forces varies, impacting \XNAME's energy consumption and maneuverability. By adjusting the output weighting of these forces, energy-saving or high-mobility movement can be achieved. Assuming \XNAME's rotational degrees of freedom on the ground only include pitch and yaw (i.e., the roll angle $\alpha=0$), and neglecting the effects of aerodynamic drag and torque, a force analysis for \XNAME~moving horizontally to the right is depicted in Fig.\ref{fig_9_10}(a). Here, $O_{3}$ is the contact point between the spherical cage and the ground, $\boldsymbol{f}$ is the ground friction force, $\boldsymbol{F}_{N}$ is the ground support force, $r$ is the radius of the spherical cage structure, and $m$ is the mass of the center of gravity tilt mechanism. Considering the resistance torque due to the cage's rotation, the torque $\boldsymbol{\tau}_{f}$ generated by the servo can be equated to the torque exerted by ground friction $\boldsymbol{f}$ at point $O_{2}$.

The Newton-Euler equations for \XNAME~in planar mode are given by
\begin{subequations}
    \begin{align}
        &M\boldsymbol{a}^{\mathcal{S}} = \boldsymbol{G}^{\mathcal{S}}+\boldsymbol{f}^{\mathcal{S}}+\boldsymbol{F}^{\mathcal{S}}+F_{N}^{\mathcal{S}}\label{eq7}\\
        &\boldsymbol{J}^{\mathcal{B}}\dot{\boldsymbol{\omega}}^{\mathcal{B}} + \boldsymbol{\omega}^{\mathcal{B}}\times \boldsymbol{J}^{\mathcal{B}}\boldsymbol{\omega}^{\mathcal{B}} = \boldsymbol{\tau} _{f}^{\mathcal{B}} + \boldsymbol{\tau} _{m}^{\mathcal{B}}+\boldsymbol{\tau} _{p}^{\mathcal{B}}\label{eq8}
    \end{align}
\end{subequations}
\noindent where $\boldsymbol{\tau} _{m}^{\mathcal{B}}=\left [ 0\;\;\;mglsin\theta\;\;\;0 \right ]^{T}$ is the torque from the center of gravity tilt mechanism at $O_{2}$. Taking $O_{3}$ as the center of rotation, the net torque is 
\begin{equation}\label{eq9}
  \tau_{all} = Frsin(\beta - \theta) + mglsin\theta  
\end{equation}

According to the relationship between the net force and the net moment, the ground friction $f$ can be expressed as 
\begin{equation}\label{eq10}
  f = \frac{1}{J_{yy}}Mr\tau_{all} - Fsin(\beta - \theta)   
\end{equation}

Note that considering the actual motion of \XNAME~in flat mode, the magnitude of $f$ in planar mode is constrained by the pitch angle $\beta$ of the brushless motor and the friction coefficient $\mu$. The range of pitch angle $\beta$ is determined by the stroke of the servo in the motor steering mechanism. The friction coefficient $\mu$ defines the maximum friction force that the ground can provide, leading to
\begin{equation}\label{eq11}
  f_{max} = \mu[Mg-Fcos(\beta - \theta)] 
\end{equation}

Equations (\ref{eq7}) and (\ref{eq8}) can be used to obtain the translational dynamics equation of mass center of \XNAME~in the $\mathcal{S}$ frame and the angular dynamics equation in the $\mathcal{B}$ frame
\begin{small}
    \begin{subequations}
        \begin{align}
            \ddot{x}&=\frac{r}{J_{yy}}C_{\psi}[FrS_{\beta - \theta} + mglS_{\theta}]\label{eq12_1}\\
            \ddot{y}&=\frac{r}{J_{yy}}S_{\psi}[FrS_{\beta - \theta} + mglS_{\theta}]\label{eq12_2}\\
            \ddot{\theta}&= -\frac{1}{J_{yy}}[(J_{xx}-J_{zz})\dot{\varphi}\dot{\psi} - mglS_{\theta}]\label{eq12_3}\\ 
            \ddot{\psi}&= \frac{1}{J_{zz}}[-\tau_{fz}C_{\theta} + \tau_{p}C_{\beta + \theta} + (J_{xx}-J_{yy})\dot{\varphi}\dot{\theta}]\label{eq12_4}
        \end{align}
    \end{subequations}
\end{small}

\subsection{Inclined Mode}

The force analysis for \XNAME~moving to the right along a slope is illustrated in Fig.\ref{fig_9_10}(b). The Newton-Euler equations in inclined mode are analogous to those in planar mode. Ignoring \XNAME's rotation on the slope, that is, $\boldsymbol{\tau} _{p}^{\mathcal{B}} = 0$. Due to the existence of the inclination angle $\gamma$ of the surface, the net torque and the ground friction force become
\begin{subequations}
    \begin{align}
        &\tau_{all}=Frcos(\beta-\theta) + mglsin(\theta-\gamma)-Mgrsin\gamma \label{eq13}\\
        &f=\frac{1}{J_{yy}}Mr\tau_{all} + Mgsin\gamma - Fsin(\beta-\theta)\label{eq14}
    \end{align}
\end{subequations}

From equations (\ref{eq13}) and (\ref{eq14}), we obtain the center of mass translational dynamics equation and the angular dynamics equation for inclined mode as
\begin{small}
    \begin{subequations}
        \begin{align}
            \ddot{x}=&-\frac{1}{J_{yy}}rC_{\gamma}C_{\psi}(MgS_{\gamma} - FrC_{\beta-\theta} + mglS_{\gamma-\theta})\label{eq15_1}\\
            \ddot{y}=&-\frac{1}{J_{yy}}rC_{\gamma}S_{\psi}(MgS_{\gamma} - FrC_{\beta-\theta} + mglS_{\gamma-\theta})\label{eq15_2}\\
            \ddot{z}=&\frac{1}{J_{yy}}rS_{\gamma}(MgS_{\gamma} - FrC_{\beta-\theta} + mglS_{\gamma-\theta})\label{eq15_3}\\
            \ddot{\theta} = &\frac{r}{J_{yy}}[FS_{\beta-\theta} - MgS_{\gamma} + \frac{Mr}{J_{yy}}(MgrS_{\gamma} - FrC_{\beta-\theta}\notag\\
            &+ mglS_{\gamma-\theta}) - (J_{xx}-J_{zz})\dot{\varphi}\dot{\psi} + mglS_{\gamma+\theta}]\label{eq15_4}
        \end{align}
    \end{subequations}
\end{small}

\section{Control and Energy Consumption}

This section establishes the appropriate controller architecture for \XNAME~based on its dynamic equations across different modes. By leveraging the adaptability of \XNAME's center of gravity height, we analyze its maneuverability and energy consumption during ground movement (including planar and inclined modes) and develop control strategies for varying operational demands.

\subsection{Aerial Mode Controller}

\begin{figure}[t]
  \centering
  \includegraphics[width=1\columnwidth]{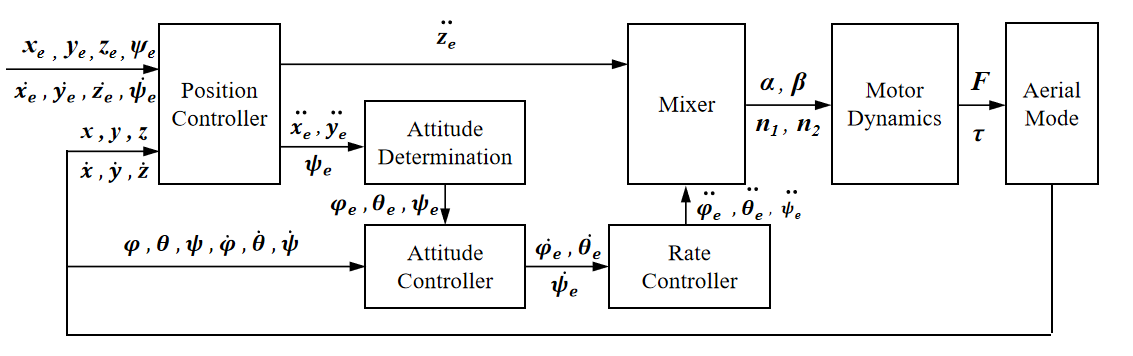}
  \caption{The architecture of aerial mode controller}
  \label{fig_11}
\end{figure}

The control system architecture for \XNAME~in aerial mode is illustrated in Fig.\ref{fig_11}. Each level of \XNAME's controller is designed as a cascade system, employing a standard PID control loop. For detailed implementation, please refer to \cite{ref10} and \cite{ref11}. The general form is represented as
\begin{equation}\label{eq17}
    u^{A} = K_{p}^{A}(\Gamma_{e} - \Gamma) + K_{i}^{A}\int (\Gamma_{e} - \Gamma)dt + K_{d}^{A}(\dot{\Gamma_{e}} - \dot{\Gamma})
\end{equation}
where $\Gamma \triangleq \left \{x, y, z, \dot{x}, \dot{y}, \dot{z}, \varphi, \theta, \psi, \dot{\varphi}, \dot{\theta}, \dot{\psi} \right \}$ and $u^{A}$ represents the controller output in aerial mode.

\subsection{Terrestrial Mode Controller}

Both planar and inclined modes belong to ground movement, thus categorized as terrestrial modes. When \XNAME~operates in terrestrial mode, it is subject to non-holonomic constraints, resulting in a notable change in degrees of freedom compared to aerial mode. Moreover, due to the unique design of \XNAME's structure, its movement on the ground can be driven by multiple sources. \XNAME~can utilize the torque generated from the center of gravity tilt mechanism for motion, as well as thrust generated from rotor tilt. These two modes of propulsion can also coexist. However, this multiplicity of movement methods complicates the control of \XNAME. Therefore, to ensure the uniqueness of the control system, it is essential to define \XNAME's movement modes in terrestrial mode.

According to the energy consumption, the movement mode is divided into energy-saving mode and high-mobility mode. The classification basis of these two ways will be given in subsection C of this section.

\begin{figure}[t]
  \centering
  \subfloat[]{\includegraphics[width=0.32\columnwidth]{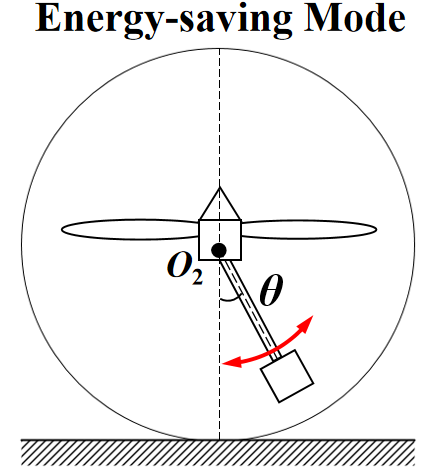}%
  \label{fig_12_1}}
  \subfloat[]{\includegraphics[width=0.32\columnwidth]{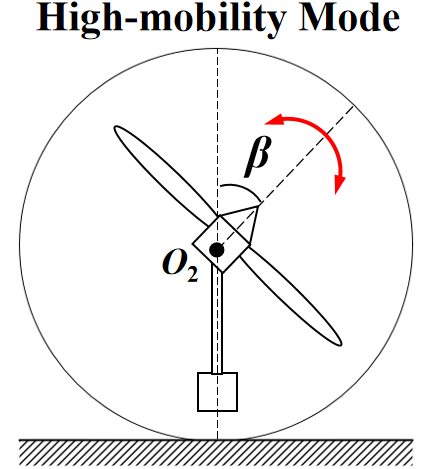}%
  \label{fig_12_2}}
  \subfloat[]{\includegraphics[width=0.32\columnwidth]{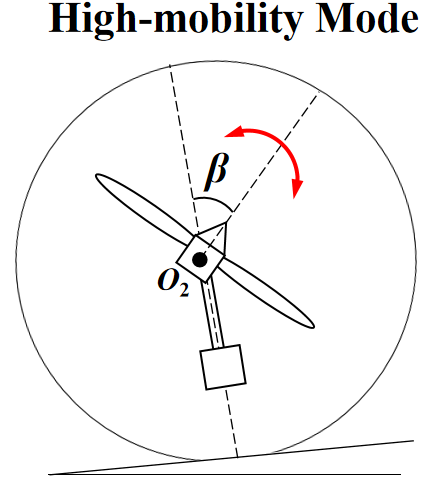}%
  \label{fig_12_3}}
  \caption{The driving methods of \XNAME~under different modes, where (a), (b) represent planar mode and (c) depicts inclined mode.}
  \label{fig_12}
\end{figure}

\textbf{Energy-saving Mode:} As shown in Fig.\ref{fig_12}(a), in this mode, the rotor plane remains parallel to the ground. \XNAME~controls its motion by tilting its center of gravity, adjusting the pitch angle $\theta$.

\textbf{High-mobility Mode:} Illustrated in Fig.\ref{fig_12}(b)(c), in this mode, the center of gravity tilt mechanism remains vertical to the ground. \XNAME~controls its motion by tilting the rotors, adjusting the motor pitch angle $\beta$.

In planar mode, \XNAME~can select between energy-saving and high-mobility modes based on the upper-level velocity expectations. In inclined mode, to ensure sufficient power, \XNAME's movement is fixed in high-mobility mode. Importantly, regardless of the selected mode or movement method, yaw control in terrestrial mode is achieved via differential speed from the rotors.

\begin{figure}[t]
  \centering
  \includegraphics[width=1\columnwidth]{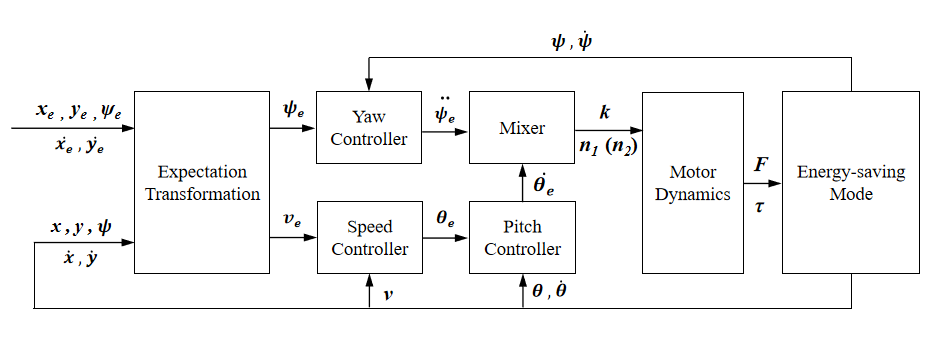}
  \caption{The architecture of energy-saving mode controller}
  \label{fig_13}
\end{figure}

The controller architecture for the energy-saving mode is illustrated in Fig.\ref{fig_13}. Due to the presence of non-holonomic constraints, the \XNAME~can only move in the heading direction on the ground. Thus, it is necessary to decompose the upper-level expectations into desired speed $v_{e}$ in the heading direction and desired yaw of the fuselage $\theta_{e}$ through a coordinate transformation module. The yaw control in energy-saving mode is similar to that in aerial mode, while speed control is achieved through center of gravity tilt. The control laws for the speed controller and the pitch controller are expressed in the following form
\begin{subequations}
    \begin{align}
        \theta_{e} &= K_{p(E)}^{v}(v_{e}-v) + K_{i(E)}^{v}\int (v_{e}-v)dt\label{eq20_1}\\
        \dot{\theta_{e}} &= K_{p(E)}^{\theta}(\theta_{e}-\theta) + K_{i(E)}^{\theta}\int (\theta_{e}-\theta)dt\label{eq20_2}
      \end{align}
\end{subequations}

Note that in energy-saving mode, the blades provide torque rather than thrust. Therefore, to minimize energy consumption, differential effects are achieved by controlling the rotation of individual blade. Additionally, in energy-saving mode, the blades must remain parallel to the ground at all times.

\begin{figure}[t] 
  \centering
  \includegraphics[width=1\columnwidth]{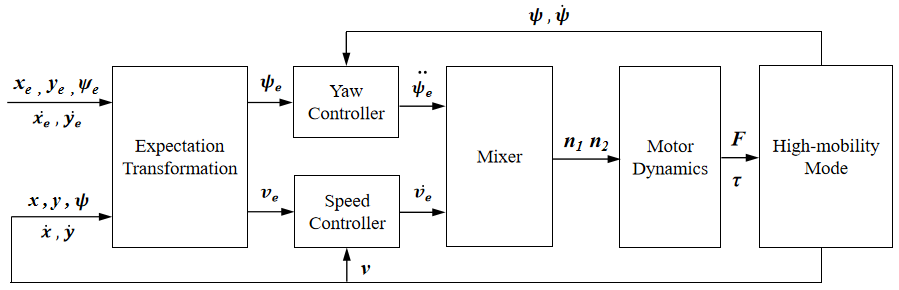}
  \caption{The architecture of high-mobility mode controller}
  \label{fig_14}
\end{figure}

The controller architecture for the high-mobility mode is shown in Fig.\ref{fig_14}. Similar to the energy-saving mode, the upper-level expectations need to be transformed into desired speed in the heading direction and desired yaw of the fuselage. However, in this case, the propulsion of the \XNAME~is provided mainly by motor thrust. In high-mobility mode, to ensure the uniqueness of the solution, the pitch angle of the servo $\beta$ must be maintained at its maximum. Due to mechanical constraints, the maximum value of $\beta$ is significantly less than 90 degrees, which prevents inadequate yaw control capabilities for the \XNAME~in high-mobility mode.

\subsection{Energy Consumption Analysis in Terrestrial Mode}

Energy efficiency is a significant consideration in vehicle design, making energy utilization efficiency an essential metric for performance evaluation. This subsection analyzes \XNAME's power consumption under different drive methods to identify the optimal movement strategy in terrestrial mode.

\begin{table*}[t]
  \caption{\textbf{Model parameters}}
  \centering
  \label{table_1}
  \renewcommand{\arraystretch}{1.5}
  \begin{tabular*}{\linewidth}{p{5.6cm}<{\centering}p{1cm}<{\centering}p{1.2cm}<{\centering}p{5.6cm}<{\centering}p{1cm}<{\centering}p{1.2cm}<{\centering}}
    \toprule[1.5pt]
    \textbf{Parameters} & \textbf{Symbol} & \textbf{Value} &
    \textbf{Parameters} & \textbf{Symbol} & \textbf{Value}\\
    \midrule
    The total mass of \XNAME & $M$ & $0.22kg$ &
    The mass of the center of gravity tilt mechanism & $m$ & $0.11kg$ \\
    Acceleration of gravity & $g$ & $9.8m/s^{2}$ &
    Ball cage radius & $r$ & $0.12m$ \\
    The distance between centroid and center & $l$ & $0.07m$ &
    Power consumption coefficient of motor & $C_{k}$ & $10.0$ \\
    Coefficient of ground static friction & $\mu$ & $0.35$ &
    Rotational inertia of \XNAME & $J_{yy}$ & $1 kg\cdot m^{2}$\\
    Maximum pitch angle of the body & $\theta_{max}$ & $15^{\circ}$ &
    Minimum pitch angle of the body & $\theta_{min}$ & $-15^{\circ}$ \\
    Maximum pitch angle of the motor & $\beta_{max}$ & $60^{\circ}$ &
    Minimum pitch angle of the motor & $\beta_{min}$ & $-60^{\circ}$ \\
    Maximum thrust of propellers & $F_{max}$ & $7.84N$ &
    Maximum thrust of propellers & $F_{min}$ & $0N$ \\
    \bottomrule[1.5pt]
  \end{tabular*}
\end{table*}

Introducing actual parameters helps provide a clearer understanding of \XNAME's power consumption composition. Table \ref{table_1} lists some parameters of \XNAME. First, we analyze \XNAME's performance in planar mode. The total power consumption of \XNAME~can be expressed as
\begin{equation}\label{eq22}
    P = P_{p} + P_{s}
\end{equation}
where $P$ denotes total power consumption, $P_{p}$ is the power consumed by the motors, and $P_{s}$ is the power consumed by the servos. For this analysis, we ignore the power consumption of the servo responsible for motor steering mechanism and focus on the servo controlling the rotation of the spherical cage. The power consumed by the motors is approximately a quadratic function of the thrust produced by the rotors. Assuming we disregard internal resistance losses, the servo power is primarily used for rotating the spherical cage and tilting the center of gravity. The formulas for calculating motor and servo power are given by
\begin{subequations}
    \begin{align}
        P_{p} &= C_{k}F^{2}\label{eq23_1}\\
        P_{s} &= \frac{1}{r}mglv_{b}sin\theta\label{eq23_2}
    \end{align}
\end{subequations}
where $C_{k}$ is the motor power coefficient and $v_{b}$ is the linear velocity at the surface of the spherical cage, indicating \XNAME's speed on the ground.

The nonlinear programming method with multi-objective constraints is used to solve the energy-optimal strategy for \XNAME~in planar mode under various acceleration expectations. The variables include the body pitch angle $\theta$, the motor pitch angle $\beta$, and the thrust $F$. The ranges of these variables are specified in Table \ref{table_1}. The nonlinear programming formulation for \XNAME~in planar mode is as follows
\begin{small}
    \begin{subequations}
        \begin{align}
            &MinP(x) = C_{k}F^{2} + \frac{1}{r}mglsin\theta \sqrt{ar}\label{eq24_1}\\
            &\frac{Mr}{J_{yy}}(FrS_{\beta-\theta} + mglS{\theta}) - Ma = 0\label{eq24_2}\\
            &F_{N} = Mg - FC_{\beta - \theta} > 0\label{eq24_3}\\
            &\frac{Mr}{J_{yy}}(FrS_{\beta-\theta} + mglS{\theta}) - FS_{\beta - \theta} - \mu F_{N} < 0\label{eq24_4}\\
            &\theta_{min} \le \theta \le \theta_{max}\label{eq24_5}\\
            &\beta_{min} \le \beta \le \beta_{max}\label{eq24_6}\\
            &F_{min} \le F \le F_{max}\label{eq24_7}
        \end{align}
    \end{subequations}
\end{small}
\!\!\!where (\ref{eq24_1}) represents the optimization objective, and the optimal energy consumption is taken as the standard. (\ref{eq24_2}) represents the acceleration expectation constraint, that is, the combined external force of \XNAME~should meet the acceleration expectation. (\ref{eq24_3}) represents the ground non-holonomic constraint that the supporting force on \XNAME~is always exist. (\ref{eq24_4}) represents the no-slip constraint, that is, the friction force required by \XNAME~cannot be greater than the maximum static friction force on the ground. (\ref{eq24_5}), (\ref{eq24_6}) and (\ref{eq24_7}) represent the hardware constraints on \XNAME.

\begin{figure}[t] 
  \centering
  \includegraphics[width=0.8\columnwidth]{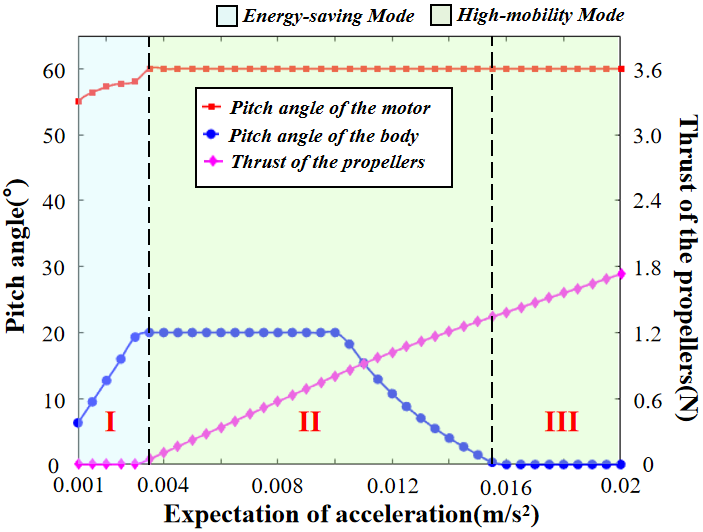}
  \caption{The variation of $\theta$, $\beta$, and $F$ in planar mode with the acceleration expectation under the optimal energy consumption.}
  \label{fig_15}
\end{figure}

\begin{figure}[t]
  \begin{minipage}[b]{\columnwidth}
    \centering
      \subfloat[]{
        \includegraphics[width=0.3\linewidth]{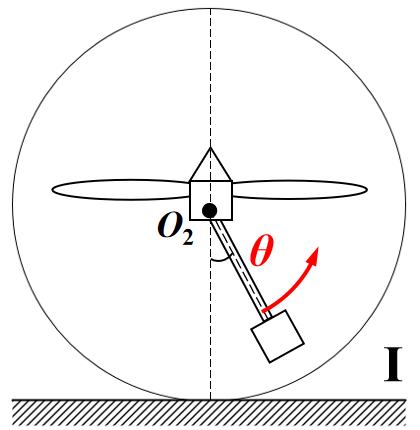}
      }
      \subfloat[]{
        \includegraphics[width=0.3\linewidth]{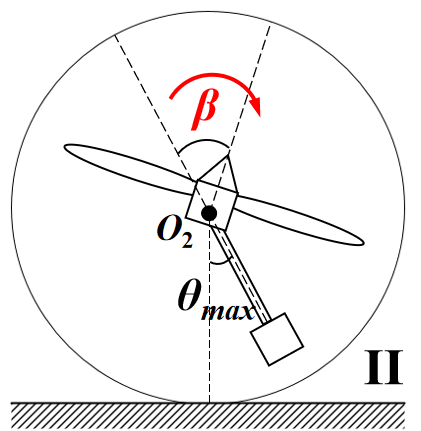}
      }
  \end{minipage}
  \begin{minipage}[b]{\columnwidth}
    \centering
      \subfloat[]{
        \includegraphics[width=0.3\linewidth]{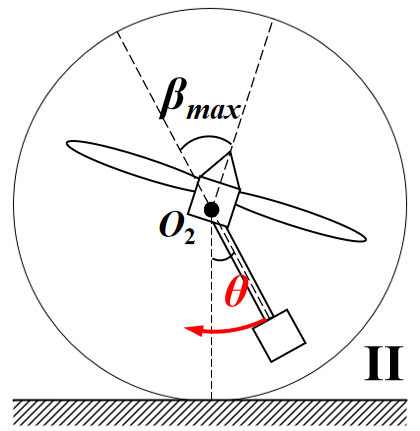}
      }
      \subfloat[]{
        \includegraphics[width=0.3\linewidth]{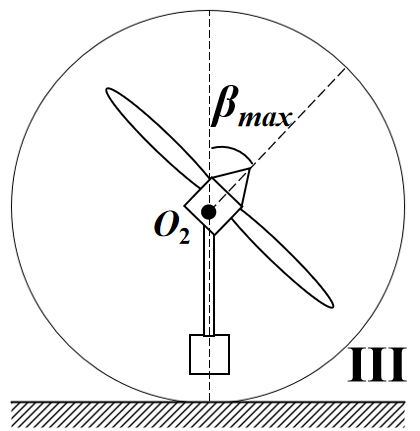}
      }
  \end{minipage}
    \caption{Process of Variation across Three Stages. (a) represents stage \Rmnum{1} where the center of gravity tilt angle $\theta$ increases. (b) indicates that in stage \Rmnum{2}, $\theta$ has reached its maximum, followed by an increase in the motor angle $\beta$. (c) indicates that in stage \Rmnum{2}, when $\beta$ reaches its maximum and the desired acceleration is still unmet, thrust is increased in the direction of motion by maintaining $\beta_{max}$ while reducing $\theta$. (d) indicates that in stage \Rmnum{3}, $\theta$ and $\beta$ are maintained at 0 and $\beta_{max}$, respectively, to satisfy the desired acceleration through increased thrust.}
    \label{fig_16}
\end{figure}

In planar mode, the variations of $\theta$, $\beta$, and $F$ under optimal energy consumption conditions with respect to the desired acceleration are illustrated in Fig.\ref{fig_15}. The horizontal axis represents the desired acceleration, while the vertical axis correspond to the pitch angle and the thrust of the propeller. The desired acceleration starts from $0.001m/s^{2}$ and varies up to $0.02m/s^{2}$. The changes in the curves can be categorized into three stages. In stage \Rmnum{1}, since the power consumption of the servo is much lower than that of the motor, the \XNAME~initially generates acceleration by changing the position of the centroid. When the centroid reaches its maximum height and the desired acceleration is still not met, \XNAME~then utilizes the motor to produce additional acceleration, leading to stage \Rmnum{2}. It is noteworthy that to maximize the horizontal component of thrust, the pitch angle $\beta$ must already be at its maximum value $\beta_{max}$ when thrust is first generated. In stage \Rmnum{3}, as the desired acceleration continues to increase, the centroid begins to descend in order to provide a greater horizontal force component. At this point, the pitch angle $\beta$ remains unchanged. The progression of the \XNAME~across these three stages is depicted in Fig.\ref{fig_16}.

Based on the results of the energy optimization, \XNAME's movement in terrestrial mode is categorized into energy-saving and high-mobility modes. In planar mode, the selection between these two modes depends on acceleration expectations. In inclined mode, due to the influence of slope angle $\gamma$, \XNAME~has limited capability to provide acceleration through center of gravity adjustments. Therefore, its movement in inclined mode is restricted to high-mobility mode.

\section{Experimental Validation}

In this section, we constructed a test environment to validate \XNAME's performance and control strategies in real-world conditions. We used UWBs to receive positional information from the vehicle and to send desired trajectories. A series of ground and aerial experiments demonstrated the effectiveness of \XNAME's design and control strategies. More detailed experimental information can be found in the supplementary video material.

\subsection{Experiments Setup}

\begin{figure*}[t]
  \begin{minipage}[b]{\textwidth}
    \centering
      \subfloat[]{
        \includegraphics[width=0.48\textwidth]{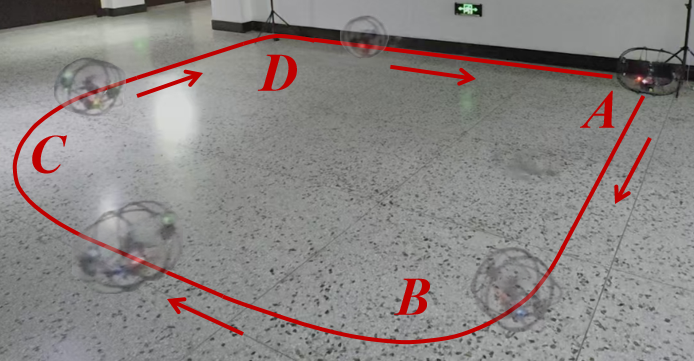}
      }
      \subfloat[]{
        \includegraphics[width=0.48\textwidth]{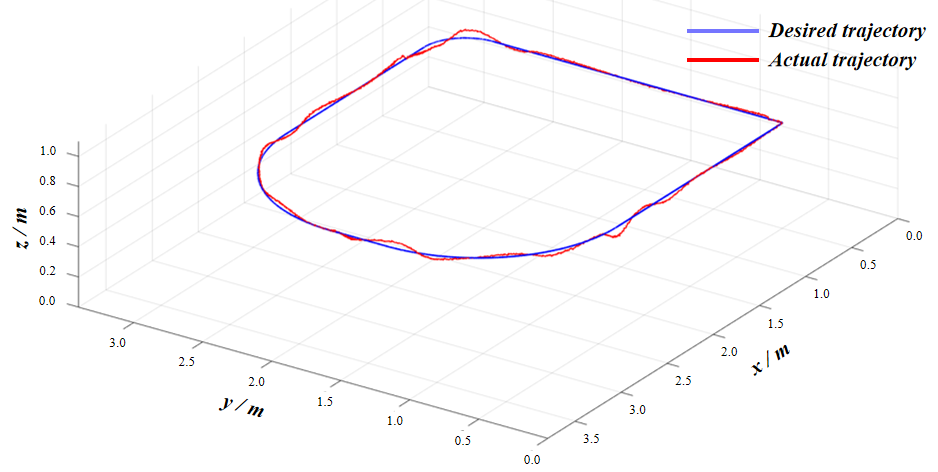}
      }
  \end{minipage}
  \begin{minipage}[b]{\textwidth}
    \centering
      \subfloat[]{
        \includegraphics[width=0.96\textwidth]{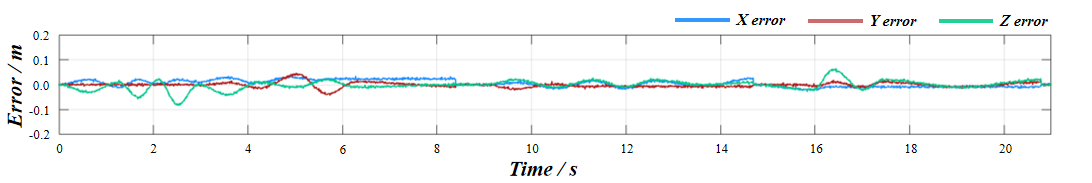}
      }
  \end{minipage}
    \caption{\XNAME's flight experiment in the test area. (a) shows the vehicle's flight trajectory starting from point A, proceeding clockwise through points B, C, and D before returning to point A. (b) compares the actual trajectory recorded during the flight with the desired trajectory. (c) depicts the changes in positional error along the $x$, $y$, and $z$ axes over time.}
    \label{fig_17}
\end{figure*}

\begin{figure}[t] 
  \centering
  \includegraphics[width=1\columnwidth]{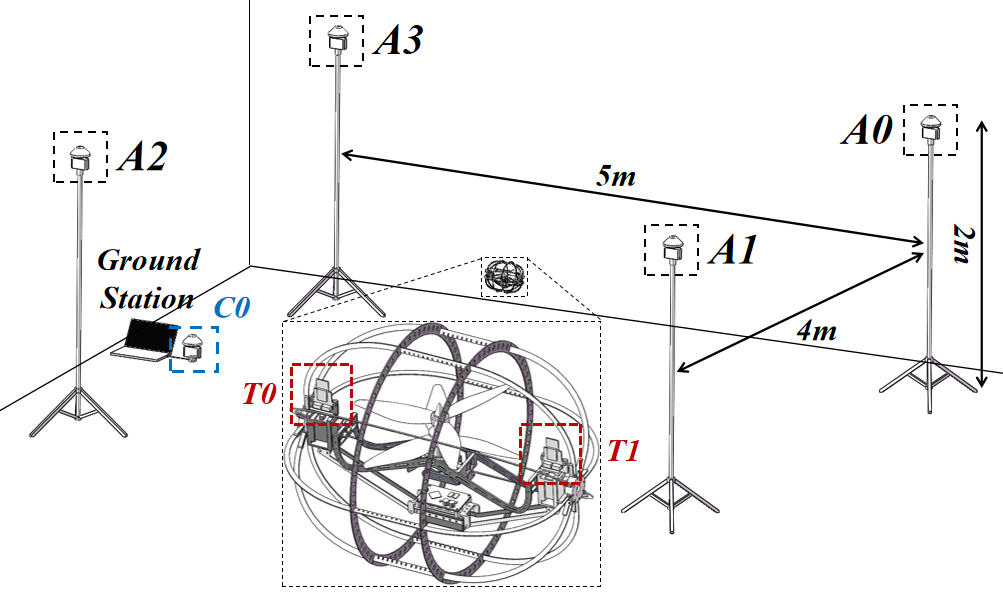}
  \caption{Indoor test environment and UWBs installation location.}
  \label{fig_18}
\end{figure}

The experiments were conducted indoors within a rectangular test area measuring $5m\times4m$. As shown in Fig.\ref{fig_18}, UWBs (A0, A1, A2, A3) were fixed at a height of $2m$ above the ground using metal tripods at each corner of the rectangle. Additionally, two miniaturized UWB tags (T1, T2) were installed within \XNAME's spherical cage. Both T0 and T1 weighed only $1.5g$ and served solely for data transmission. The devices A0-A3 were capable of positional calculations, while an additional UWB (C0) functioned as a control console connected to the ground station, which gathered positional data and sent target positions to \XNAME.

\subsection{Aerial Experiment}

We first validated \XNAME's motion performance and control strategies in aerial mode. To test \XNAME's maneuverability in the air, we designed two curves with different curvatures. As shown in Fig.\ref{fig_17}(a), the arcs at points B and C have the same curvature with a radius of $1m$, while point D features a smaller radius of only $0.3m$, approximating a right-angle turn. The comparison between the expected and actual trajectories reveals that \XNAME's body attitude experiences greater fluctuations during arc movements compared to linear movements. This phenomenon arises from two factors: firstly, the characteristics of the coaxial tilt-rotor control, where yaw is achieved through differential speeds from the rotors, causing fluctuations in thrust due to variations in rotor speeds; secondly, the non-linear relationship between thrust and control signals, resulting in differing control precision across various operating ranges. Despite these effects, Fig.\ref{fig_17}(c) shows that the positional errors in the $x$, $y$, and $z$ axes generally remain within $\pm5cm$, with overall deviations not exceeding $\pm10cm$. Thus, \XNAME's aerial motion performance meets expectations, confirming the effectiveness of the control strategy in aerial mode.

\subsection{Ground Experiment}

\begin{figure*}[t]
  \begin{minipage}[b]{\textwidth}
    \centering
      \subfloat[]{
        \includegraphics[width=0.48\textwidth]{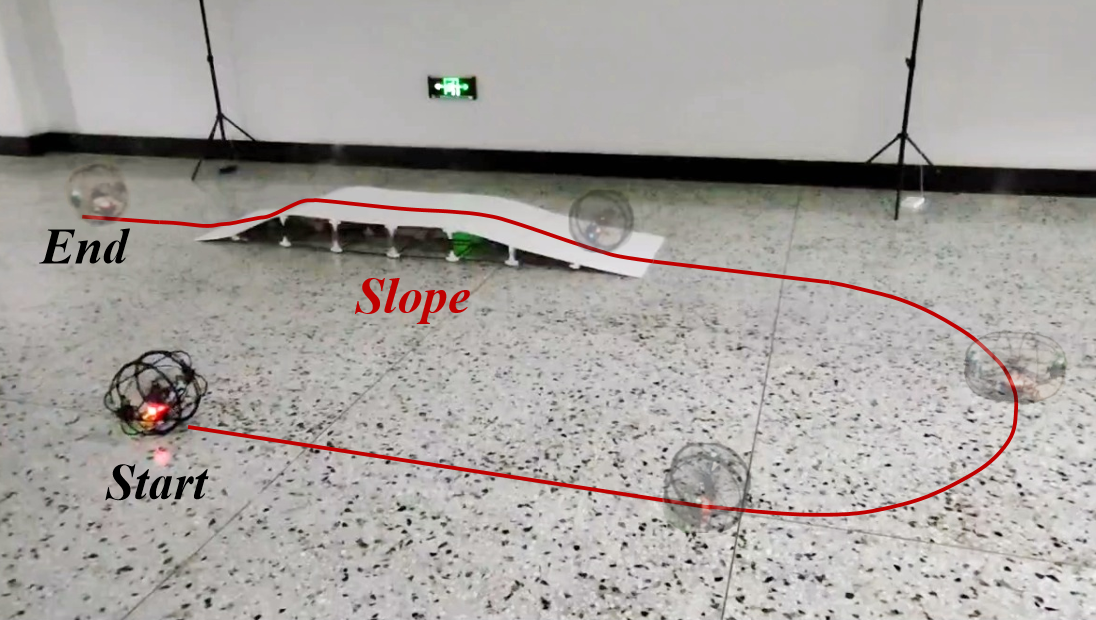}
      }
      \subfloat[]{
        \includegraphics[width=0.48\textwidth]{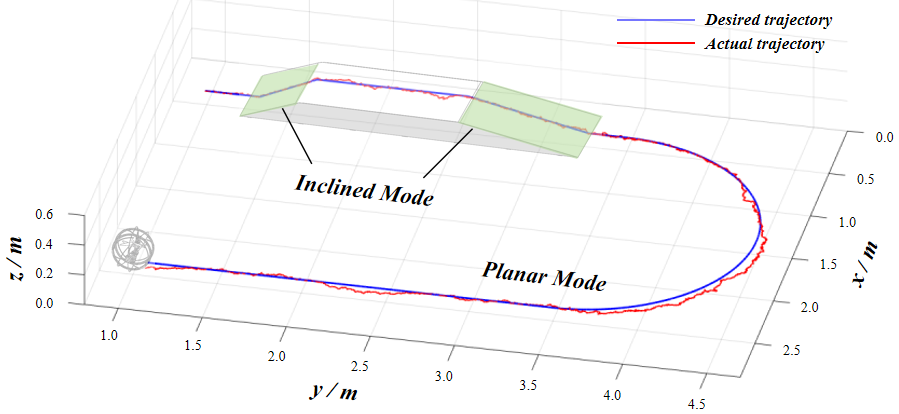}
      }
  \end{minipage}
  \begin{minipage}[b]{\textwidth}
    \centering
      \subfloat[]{
        \includegraphics[width=0.96\textwidth]{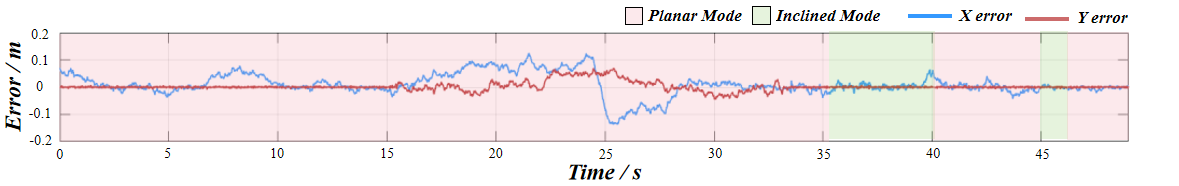}
      }
  \end{minipage}
    \caption{Ground experiments in the test site. (a) The ground motion trajectory of \XNAME, starting from ``$Start$'', passing through straight and semi-circular turns, and reaching ``$End$'' through a slope. (b) Comparison between the actual recorded trajectory of \XNAME~and the expected trajectory during the ground experiment. (c) Curve of the error of \XNAME~in $x$ and $y$ directions as a function of time during the experiment.}
    \label{fig_19}
\end{figure*}

\begin{figure}[t] 
  \centering
  \includegraphics[width=1\columnwidth]{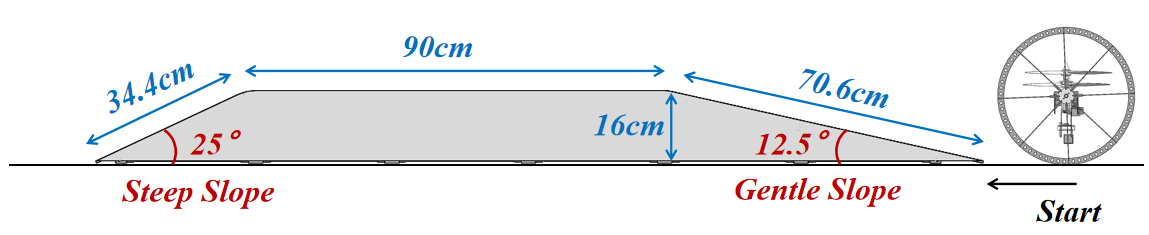}
  \caption{Slope parameters in ground experiment.}
  \label{fig_20}
\end{figure}

Subsequently, we tested \XNAME's ground performance and control efficacy. As depicted in Fig.\ref{fig_19}, \XNAME's movement includes both planar and inclined modes, with the desired path designed to initiate from the ``Start'', move in a straight line for $2.5m$, navigate a semicircular arc with a radius of $1m$, and finally traverse the slope illustrated in Fig.\ref{fig_20} to reach the ``End''.

The slope is designed for \XNAME~to enter from the gentle side and exit from the steep side, with a height of $16cm$. The gentle side has an inclination angle of $12.5^{\circ}$ and a length of $70.6cm$, while the steep side has an inclination angle of $25^{\circ}$ and a length of $34.4cm$. The top surface of the slope is parallel to the ground, with a total length of $90cm$. The purpose of this design is to test and verify \XNAME's capability in high-mobility mode while ascending, as well as its stability during descent. The supplementary video demonstrates that while \XNAME~moves slowly uphill (due to the steep gradient), it successfully and smoothly reaches the top. During descent, \XNAME~maintains a stable attitude without losing control or rolling over due to excessive speed.

Fig.\ref{fig_19}(c) illustrates the variations in error over time during the experiment. Starting from the beginning, \XNAME's initial position deviated $5cm$ from the expected location, prompting it to gradually align with the desired trajectory. Upon entering the semicircular expected path, \XNAME~experienced delayed control during turns due to the significant ground friction moment (with a static friction coefficient of approximately $0.26$). As shown in Fig.\ref{fig_19}(c), at $13s$, \XNAME~entered the arc, where the deviations in the $x$ and $y$ directions increased continuously. This situation corresponds to the observed deviation from the expected trajectory during the arc entry in Fig.\ref{fig_19}(b). Subsequently, by increasing the controller output, \XNAME~returned to the expected trajectory and completed the arc movement. During slope traversal, \XNAME's attitude control remained notably stable. The error within the green region in Fig.\ref{fig_19}(b) and (c) fluctuated within $\pm2cm$. Overall, \XNAME's ground motion performance meets expectations, further validating the effectiveness of the control strategy in terrestrial mode (including both planar and inclined modes).

\section{Conclusion}

This paper investigates the design, modeling, and control of \XNAME, a hybrid terrestrial/aerial coaxial tilt-rotor vehicle. An innovative structure featuring an embedded coaxial tilt-rotor configuration encased in an external spherical cage is proposed. The dynamics modeling and controller design for this structure have been accomplished for aerial, planar, and inclined modes. The design enables flexible movement in both terrestrial and aerial modes through tilt mechanism and thrust-driven propulsion. In terrestrial mode, two movement patterns for \XNAME~are designed from the perspective of energy consumption optimization, and the output sequence of various actuators is planned for different states. Experimental results validate the movement performance and effectiveness of the control strategies of \XNAME. Future work will involve equipping \XNAME~with LiDAR, high-performance computing units, and other technologies to explore navigation planning and higher-level decision-making challenges.


\bibliographystyle{IEEEtran}
\bibliography{lookup.bib}

\vfill

\end{document}